\documentclass[11pt]{article}

\usepackage[final]{acl}

\usepackage{times}
\usepackage{latexsym}
\usepackage{booktabs}

\usepackage[T1]{fontenc}

\usepackage[utf8]{inputenc}
\usepackage{csquotes}
\usepackage{microtype}

\usepackage{inconsolata}
\usepackage{subcaption}

\usepackage{graphicx}

\title{Can Continual Pre-training Bridge the Performance Gap between General-purpose and Specialized Language Models in the Medical Domain?}

\author{ \bf
Niclas Doll$^{1,2}$,
Jasper Schulze Buschhoff$^{1}$,
Shalaka Satheesh$^{1}$, \\
\bf
Hammam Abdelwahab$^{1}$,
Héctor Allende-Cid$^{1,2}$,
Katrin Klug$^{1}$ \\ \\
$^1$Fraunhofer IAIS, $^2$Lamarr Institute \\
\{niclas.doll, johann.jasper.schulze.buschoff, shalaka.satheesh\}@iais.fraunhofer.de
}

\begin{document}

\maketitle

\begin{abstract}
This paper narrows the performance gap between small, specialized models and significantly larger general-purpose models through domain adaptation via continual pre-training and merging. We address the scarcity of specialized non-English data by constructing a high-quality German medical corpus (\textit{FineMed-de}) from \textit{FineWeb2}. This corpus is used to continually pre-train and merge three well-known LLMs (ranging from 7B to 24B parameters), creating the \textit{DeFineMed} model family. A comprehensive evaluation confirms that specialization dramatically enhances 7B model performance on German medical benchmarks. Furthermore, the pairwise win-rate analysis of the \textit{Qwen2.5}-based models demonstrates an approximately $3.5$-fold increase in the win-rate against the much larger \textit{Mistral-Small-24B-Instruct} through domain adaptation. This evidence positions specialized $7B$ models as a competitive, resource-efficient solution for complex medical instruction-following tasks. While model merging successfully restores instruction-following abilities, a subsequent failure mode analysis reveals inherent trade-offs, including the introduction of language mixing and increased verbosity, highlighting the need for more targeted fine-tuning in future work. This research provides a robust, compliant methodology for developing specialized LLMs, serving as the foundation for practical use in German-speaking healthcare contexts.
\end{abstract}

\section{Introduction}

Large language models (LLMs) have demonstrated transformative potential across various domains, including healthcare \citep{gerd23, zhang-etal-2024-comprehensive-survey}. Their ability to process and generate human-like text enables applications such as diagnostics, personalized treatment plans, and efficient information retrieval \citep{meyer2024}. However, significant gaps remain in the integration of LLMs into clinical workflows, where general-purpose models often fail to capture domain-specific knowledge and terminology with sufficient accuracy \citep{klug2024, ALONSO2024}. 

The effective application of LLMs in the medical field faces two significant, interconnected challenges. First, stringent data protection regulations necessitate on-premise solutions, making the use of large, API-based LLM services impractical and favoring smaller, computationally efficient models \citep{belcak2025}. Second, these smaller models, while being regulatory compliant, struggle to capture the complex, nuanced medical terminology due to a scarcity of high-quality, domain-specific datasets, a problem acutely felt in non-English languages \citep{german_mimic}. This creates a critical trade-off: regulatory constraints demand small models, underscoring the necessity of a targeted, specialized knowledge base to achieve competitive clinical performance.

This work directly addresses this challenge by investigating the core research question: \textit{Can domain-adaptation of $7B$ language models, achieved via continual pre-training and merging, close the performance gap sufficiently to compete with significantly larger, general-purpose models on complex medical tasks?} Beyond technical constraints, this study aims to quantify how effectively domain knowledge, gained without massive retraining, can be internalized to transform resource-efficient models into viable clinical tools.

\begin{figure*}[t]
    \centering
    \includegraphics[width=\linewidth]{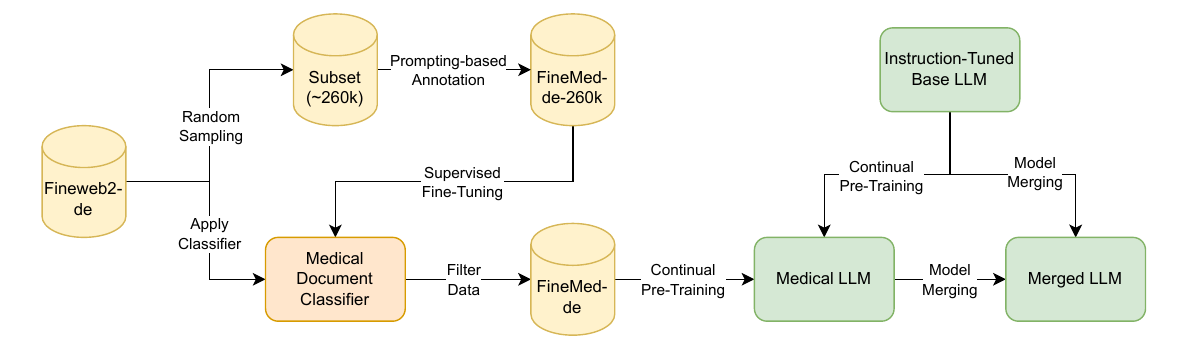}
    \caption{High-level illustration of the Data Filtering and Model Adaptation workflow: a subset of the German \textit{FineWeb2} dataset is annotated into medical and non-medical documents. The resulting annotated dataset, \textit{FineMed-de-260k}, is used to train a classifier, which is then applied to the full German \textit{FineWeb2} split, producing the \textit{FineMed-de} dataset. This dataset is subsequently used for continual pre-training - resulting in a Medical LLM - which is then merged with the initial instruction-tuned checkpoint.}
    \label{fig:workflow}
\end{figure*}

This paper directly addresses these challenges by developing a specialized German medical LLM through a systematic approach to dataset creation and model adaptation, as illustrated in Figure \ref{fig:workflow}. Our key contributions are as follows:
\begin{itemize}
\item A robust and scalable filtering methodology for creating high-quality, domain-specific datasets from general-purpose corpora. This methodology combines LLM-based annotation with machine learning techniques and is applicable to various domains and languages.
\item A comprehensive evaluation, utilizing knowledge-intensive benchmarks (\textit{MMLU-de} and \textit{MedQA-de}) and a pairwise win-rate analysis, demonstrating that specialized $7B$ models drastically close the performance gap toward $24B$ models. This validates the competitive viability of smaller, resource-efficient LLMs, while a simultaneous failure mode analysis highlights the critical trade-offs inherent to the specialization methodology.
\end{itemize}

The remainder of this paper is structured as follows: Section \ref{sec:related} provides an overview of the state of the art. Section \ref{sec:medical_fineweb} outlines the construction of the German medical pre-training dataset. Further, Section \ref{sec:model_adaptation} describes the model adaptation techniques employed. The evaluation methodology and results are presented in Section \ref{sec:evaluation}, followed by a discussion of the observed results in Section \ref{sec:discussion}. Finally, Section \ref{sec:conclusion} concludes the paper with a summary of our findings and potential future directions.

\section{Related Work}
\label{sec:related}
This work intersects several key areas of research in medical LLM development, including model adaptation through continual pre-training and model merging techniques, as well as dataset curation through document filtering. To provide a comprehensive context, we review relevant literature in each of these domains, highlighting advancements and methodologies that align with our approach.

\subsection{Model Adaptation through Continual Pre-training}

Recent studies have explored continual pre-training as a means to adapt general-purpose LLMs to specialized domains, but results remain inconsistent. \citet{oncel2024} showed that additional pre-training can fail, or even harm performance, when domain data diverges from the model’s original distribution, underscoring the importance of corpus quality and alignment. In contrast, research in the legal and financial domains \citep{niyogi2024, siriwardhana2024} suggests that combining continual pre-training with model merging can yield competitive improvements while preserving general reasoning capabilities. Similarly, \citet{Yang2024b} showed that targeted continual pre-training with efficient adaptation strategies enhances domain fluency, suggesting that success depends less on scale and more on data selection and adaptation design.

\subsection{Medical Large Language Models}

Recent progress in Medical LLMs has been achieved through pre-training \citep{Luo_2022, Peng_2023}, fine-tuning \citep{singhal2025, toma2023, han2023, zheng2025}, and prompting \citep{nori2023, liu2023}, yielding models that match or surpass human experts on medical \textit{QA} tasks \citep{singhal2025, nori2023}.

\citet{labrak2024biomistralcollectionopensourcepretrained} advanced this line with \textit{BioMistral}, a continually pre-trained biomedical model based on Mistral \citep{jiang2023mistral7b} and PubMed-Central data \citep{pmc_open_access_subset}, outperforming open-source baselines. Our work fundamentally differs from \textit{BioMistral} not only in targeting the German language but in its primary research objective. Where \textit{BioMistral} aims for performance improvements on established, knowledge-intensive benchmarks, our study investigates the competitive viability of specialized small LMs against significantly larger general-purpose models and extends the evaluation beyond knowledge-intensive benchmarks.

More recently, \citet{zheng2025} introduced \textit{Apollo-2}, a multilingual medical LLM family covering 50 languages, focusing on instruction-tuning and modular routing rather than continual pre-training for domain adaptation.

\subsection{Merging Large Language Models}

Model merging provides an efficient alternative to computationally expensive fine-tuning and ensembling by combining pre-trained LLMs fine-tuned on specialized tasks. The survey by \citet{yang2024modelmergingllmsmllms} presents a taxonomy of model merging approaches, their applications in various domain subfields, and their challenges. Additionally, \citet{nobari2025} propose \textit{Activation-Informed Merging}, leveraging activation-space information to improve the robustness and efficiency of merging, showing up to a $40\%$ performance gain across benchmarks.

Alternative strategies further explore computational efficiency and generalization. \citet{gupta2024model} introduce a task-vector-based approach using LoRA-derived representations and geometric median aggregation, enabling effective merging with reduced computational costs. Meanwhile, \citet{yadav2024} conducted large-scale experiments, merging models up to 64B parameters, showing that merging enhances zero-shot generalization and is particularly effective when starting from strong base models. Their findings suggest that larger models facilitate more successful merging, and different merging strategies perform similarly at scale. Collectively, these works highlight the potential of model merging for creating high-performing, resource-efficient LLMs.

In this work, we employ Spherical Linear Interpolation (SLERP) \citep{shoemake-1985-slerp} for model merging, a technique shown by \citet{goddard-etal-2024-arcees} to achieve the best performance in the medical domain.

\section{Medical Filtering Pipeline}
\label{sec:medical_fineweb}

This section details the creation of the \textit{FineMed-de} corpus. The high level process is illustrated in Figure \ref{fig:workflow}. Unlike existing curation approaches that rely solely on either classical ML or LLMs, we apply a hybrid method. By using LLMs to generate high-quality labels and classical ML techniques to scale the filtering process to millions of documents, our approach aims to achieve both quality and efficiency in medical dataset curation.

\subsection{Source Data}
\label{sec:medical_fineweb:source_data}
The source dataset utilized in this study is \textit{FineWeb2} \citep{penedo2024fineweb2}, a web-crawled dataset designed to improve accessibility for training large language models. \textit{FineWeb2} is built upon \textit{CommonCrawl}\footnote{\url{https://commoncrawl.org/}}, encompassing a wide variety of web content including forum posts, blog articles, and news articles. The original \textit{FineWeb2} dataset is partitioned by language using the \textit{GlotLID} language classifier \citep{kargaran2023glotlid}, with the German subset being a substantial portion, ranking as the third largest at $640GB$. \textit{FineWeb2} is licensed under the \textit{Open Data Commons Attribution License (ODC-By) v1.0}, allowing for its open use and distribution. Our objective is to filter this subset to extract exclusively medical documents for pre-training specialized German medical LLMs.

\subsection{Medical Document Classifier}
\label{sec:classifier}

To create a robust medical pre-training corpus, we first developed a medical document classifier. This involved constructing a supervised dataset by sampling approximately $260k$ documents from the German \textit{FineWeb2} dataset. We then employed the \textit{Mixtral-8x7B-Instruct-v0.1} \citep{jiang2024mixtralexperts} model to partition these documents into medical and non-medical categories using a zero-shot prompting approach. At the time of our study, this model offered a combination of strong multilingual performance and computational efficiency, making it a practical choice for our classification task. The prompt used is provided in Appendix \ref{sec:med_pre_cls}. The output of the LLM was validated through manual inspection of $100$ random samples by three human annotators, yielding an F1 score of $91.1 \pm 2.5$. The manual inspection process and further evaluation of the LLM's performance is detailed in Appendix \ref{sec:med_pre_cls}. The resulting labeled dataset, named \textit{FineMed-de-260k}, was then randomly split into training and testing sets, as detailed in Table \ref{tab:classifier_training_data}.

To achieve a cost-effective and scalable solution for classifying the entire pre-training corpus, we train a significantly smaller and more efficient classifier based on this annotated dataset. Specifically, we fine-tune a $279M$ parameter \textit{xlm-roberta} model \citep{conneau2020unsupervisedcrosslingualrepresentationlearning} as a medical document classifier, achieving a precision of $0.95$ and recall of $0.8$ on the test set. Details of the training procedure can be found in Appendix \ref{app:med_cls}.

\begin{table}[t]
    \centering
    \begin{minipage}{0.48\textwidth}
        \centering
        \begin{tabular}{l|rr|rr}
            & \multicolumn{2}{c|}{\textbf{\#Documents}} & \multicolumn{2}{c}{\textbf{\#Words}} \\
            & \textbf{Med} & \textbf{Other} & \textbf{Med} & \textbf{Other} \\
            \hline
            Test & $4.9k$ & $21.5k$ & $8.7M$ & $20.3M$ \\
            Train & $44.1k$ & $193.5k$ & $81.9M$ & $181.8M$ \\
            Total & $49.0k$ & $215.0k$ & $90.6M$ & $202.1M$ \\
        \end{tabular}
        \caption{Label distribution between medical (Med) and non-medical (Other) domains of the \textit{FineMed-de-260k} dataset in terms of number of documents and words.}
        \label{tab:classifier_training_data}
    \end{minipage}\hfill
    \begin{minipage}{0.48\textwidth}
        \centering
        \begin{tabular}{c|cc}
            \textbf{Dataset} & \textbf{\#Documents} & \textbf{\#Words} \\
            \hline
            FineMed-de & $7.3M$ & $5.1B$ \\
            FineWeb2-de & $427.7M$ & $234.8B$ \\
        \end{tabular}
        \caption{Dataset size of \textit{FineMed-de} compared to the original \textit{FineWeb2-de} in number of documents and number of words.}
        \label{tab:fineweb_statistics}
    \end{minipage}
\end{table}

\subsection{Medical Pre-Training Corpus}
\label{sec:finemed_data}

The application of the classifier to the entire German \textit{FineWeb2} dataset required approximately $400$ GPU hours, distributed across 8 A100 (40GB) GPUs on the Karolina cluster.\footnote{\url{https://www.it4i.cz/en/infrastructure/karolina}} The resulting dataset, which we named \textit{FineMed-de}, contains roughly $7.3$ million documents. The statistics of the resulting dataset are presented in Table \ref{tab:fineweb_statistics}.

\section{Model Adaptation}
\label{sec:model_adaptation}

To create specialized German medical LLMs, we select three source models for adaptation via continual pre-training and model merging. We name this family of adapted models \textit{DeFineMed}, a combination of "De" for German and \textit{FineMed}, our dataset used for continual pre-training. The decision to start from instruction-tuned models for our adaptation process is motivated by both methodological precedent and empirical evidence. Following the setup of the \textit{BioMistral} model family \citep{labrak2024biomistralcollectionopensourcepretrained}, which also relies on instruction-tuned checkpoints as the basis for continual pre-training and model merging, we adopt a similar strategy to ensure alignment with established practices in the domain. 
At the same time, we acknowledge that the optimal sequencing of instruction tuning and continual pre-training remains an open research question, with recent studies reporting conflicting findings on whether instruction tuning should precede or follow domain adaptation \citep{jindal2024balancing, jiang2024instruction}.

\subsection{Source Models}
\label{sec:source_models}

The foundation of our model adaptation process is the selection of appropriate source models. We prioritize models with strong multilingual capabilities to establish a robust base for understanding and generating text in our target language. Specifically, we focus on \textit{Qwen2.5-7B-Instruct} and \textit{Mistral-7B-Instruct}, both of which offer a balance between performance and computational efficiency within the $7B$ parameter range.
To investigate how our process scales with model size, we also included \textit{Mistral-Small-24B-Instruct} with 23.6B parameters. The complete list of models used in this work is provided in the Table \ref{tab:source_models}. 
\begin{table*}[]
    \centering
    \begin{tabular}{l|cc|cc}
        \textbf{Model} & \textbf{Parameters} & \textbf{Vocab Size} & \textbf{License} & \textbf{Source} \\
        \hline
        Mistral-7B-Instruct-v0.3 & $7.25B$ & $32k$ & \textit{Apache-2.0} & \citep{jiang2023mistral7b} \\
        Qwen2.5-7B-Instruct & $7.62B$ & $152k$ & \textit{Apache-2.0} & \citep{qwen2025qwen25technicalreport} \\
        Mistral-Small-24B-Instruct & $23.6B$ & $131k$ & \textit{Apache-2.0} & \citep{mistral_small_3} \\
    \end{tabular}
    \caption{Source model information, including model size, vocabulary size, license citations.}
    \label{tab:source_models}
\end{table*}

\subsection{Continual Pre-training}
\label{sec:pretraining}

To adapt the source models to the medical domain, we perform continual pre-training using the \textit{FineMed-de} dataset from Section \ref{sec:medical_fineweb}. We leverage the Hugging Face \textit{Transformers} library \citep{wolf-etal-2020-transformers} alongside the \textit{Accelerate} library \citep{accelerate} for efficient distributed training. To optimize both memory usage and computational performance, we incorporate several advanced techniques such as \textit{Fully Sharded Data Parallelism} (FSDP) \citep{zhao2023pytorchfsdpexperiencesscaling}, \textit{Flash Attention} \citep{dao2023flashattention2fasterattentionbetter}, \textit{Activation Checkpointing}, \textit{Sequence Packing} \citep{ding2024fewer}, and mixed-precision training with \textit{bfloat16}. We train all models for a total of two epochs using the \textit{AdamW} optimizer \citep{loshchilov2019decoupledweightdecayregularization} with a linear learning rate decay. In order to address training instabilities, we implement an extended warmup phase of $500$ steps. A detailed description of the pre-training procedure is given in Appendix \ref{app:training-details}. 

\subsection{Model Merging}
\label{sec:model_merging}

Inspired by the observations from \citet{labrak2024biomistralcollectionopensourcepretrained}, where merging models led to improved performance in various tasks, we employ model merging after training.
Following continual pre-training, we apply model merging to mitigate catastrophic forgetting and restore instruction-following abilities. By merging it with its original instruction-tuned version we expect to recover the instruction-following ability \citep{yang2024modelmergingllmsmllms}.
Each continually pre-trained model is merged with its corresponding instruction-tuned base model using the Mergekit framework \citep{goddard-etal-2024-arcees}. We employ SLERP \citep{shoemake-1985-slerp}, which has been shown to outperform alternative strategies in the medical domain \citep{goddard-etal-2024-arcees}, following the layer-wise, component-specific interpolation schedule proposed by \citet{Lu2025finetuning}. The full merging configuration is provided in Appendix \ref{app:merge-config}. This approach provides an efficient means of preserving generalization capabilities without additional fine-tuning.
\begin{table*}[t]
    \centering
    \resizebox{\textwidth}{!}{
    \begin{tabular}{l|cccc|c}
    \textbf{Model} & \textbf{Anatomy} & \textbf{Clinical Knowledge} & \textbf{College Medicine} & \textbf{MedQA-de} & \textbf{Average} \\ 
    \hline
    BioMistral-7B (baseline) & 44.44 ± 4.29 & 52.08 ± 3.07 & 40.46 ± 3.74 & 37.20 ± 2.16 & 43.55 ± 1.70 \\
	BioMistral-7B-SLERP (baseline) & 45.93 ± 4.30 &58.49 ± 3.03 &50.87 ± 3.81 &37.60 ± 2.17 &48.22 ± 1.71 \\ 
    \hline
	Mistral-7B-Instruct-v0.3 &42.22 ± 4.27 &61.13 ± 3.00 &53.18 ± 3.80 &42.40 ± 2.21 &49.73 ± 1.71 \\ 
	DeFineMed-Mistral-7B &\underline{57.78 ± 4.27} &65.28 ± 2.93 &55.49 ± 3.79 &46.80 ± 2.23 &56.34 ± 1.70 \\ 
	DeFineMed-Mistral-7B-SLERP &53.33 ± 4.31 &\underline{65.66 ± 2.92} &\underline{56.65 ± 3.78} &\underline{50.20 ± 2.24} &\underline{56.46 ± 1.70} \\
    \hline
	Qwen2.5-7B-Instruct &54.07 ± 4.30 &69.06 ± 2.85 &63.01 ± 3.68 &50.20 ± 2.24 &59.08 ± 1.68 \\ 
	DeFineMed-Qwen2.5-7B &62.22 ± 4.19 &\underline{76.60 ± 2.61} &\underline{68.21 ± 3.55} &52.60 ± 2.24 &64.91 ± 1.62 \\ 
	DeFineMed-Qwen2.5-7B-SLERP &\underline{65.19 ± 4.12} &75.85 ± 2.63 &65.90 ± 3.61 &\underline{54.20 ± 2.23} &\underline{65.28 ± 1.62} \\ 
    \hline
	Mistral-Small-24B-Instruct &66.67 ± 4.07 &78.87 ± 2.51 &73.41 ± 3.37 &68.80 ± 2.07 &71.94 ± 1.55 \\ 
	DeFineMed-Mistral-Small-24B &68.15 ± 4.02 &79.62 ± 2.48 &77.46 ± 3.19 &68.80 ± 2.07 &73.51 ± 1.52 \\ 
	DeFineMed-Mistral-Small-24B-SLERP &\underline{68.89 ± 4.00} &\underline{82.64 ± 2.33} &\underline{78.03 ± 3.16} &\underline{70.20 ± 2.05} &\underline{74.94 ± 1.49} \\ 
    \end{tabular}
    }
    \caption{Model performance on different German benchmarks in one-shot setting. The average accuracy is the unweighted mean of accuracies, whereas the standard error of the average is the square root of the unweighted mean of variances. Best score per model family is \underline{underlined}.}
    \label{tab:model_performance}
\end{table*}

\section{Evaluation}
\label{sec:evaluation}

This chapter presents a comprehensive evaluation of the model checkpoints across three dimensions: benchmark performance, pairwise competitive analysis, and quantitative failure mode assessment. Our analysis systematically investigates the impact of continual pre-training and model merging on both domain-specific accuracy and output quality.

We use \textit{BioMistral} as a comparative baseline throughout our evaluation due to its similar domain adaptation technique. \textit{Apollo-2} is excluded as a direct baseline due to its focus on instruction-tuning as its primary specialization mechanism.

\subsection{Knowledge-Intensive Benchmark Evaluation}
\label{sec:results}

We evaluate model performance across three checkpoints for each selected source model—the initial instruction-tuned model, the continually pre-trained model, and the final merged model. Performance is assessed using the LM Evaluation Harness \citep{eval-harness, biderman2024lmevaluation} on two established medical benchmarks: (1) \textit{MMMLU}, specifically focusing on the German medical tasks (Anatomy, Clinical Knowledge, and College Medicine), and (2) \textit{MedQA-de}, a machine-translated German version of the \textit{MedQA} dataset \citep{medqa} consisting of 500 medical exam questions.
To ensure that domain adaptation does not compromise general reasoning or linguistic capabilities, we additionally evaluate all models on a set of general knowledge benchmarks, with the corresponding results reported in Appendix \ref{sec:non_medical_eval} and in Table \ref{tab:model_performance}.

Base Models achieve very different performance across the benchmarks. \textit{Qwen2.5-7B-Instruct} outperforms the \textit{Mistral-7B-Instruct} model by $9.34\%$ on average across all benchmarks, likely due to more modern data pipelines and pre-training methodologies. Notably, the \textit{Mistral-Small-24B-Instruct} model achieves the best overall results, demonstrating a significant increase in performance, particularly on \textit{MedQA-de}, where it outperforms the second-best model by $18.60\%$. We observed that even our weakest base model outperforms the \textit{BioMistral} baseline on three out of four benchmarks. This is likely due to the primarily English corpus used to adapt the model.

Continual Pre-Training on domain-specific data consistently improves performance across multilingual models. \textit{DeFineMed-Mistral-7B} exhibits the strongest relative performance increase from continual pre-training, with an average improvement of $6.61\%$.
\textit{DeFineMed-Mistral-Small-24B} also benefits from continual pre-training, though to a lesser extent than smaller models, with an average gain of $1.57\%$. The impact of using domain-specific data is validated through an ablation study designed to isolate and quantify the benefit of our medical corpus compared to an unfiltered baseline, as detailed in Section \ref{sec:ablation-domain-specific-filtering}.

Model Merging shows mixed results. While it consistently improved performance on the \textit{MedQA} benchmark and showed consistent, though in some cases marginal, gains for the \textit{DeFineMed-Mistral-Small-24B-SLERP} model, other combinations of models and benchmarks did not reflect a clear trend. For instance, the performance of \textit{DeFineMed-Qwen2.5-7B} on the \textit{Clinical Knowledge} and \textit{College Medicine} benchmarks dropped after merging. On average, model merging slightly improved performance across the benchmarks, offering the advantage of partially restoring instruction-following abilities that were lost during continual pre-training.

In terms of overall performance, \textit{DeFineMed-Mistral-Small-24B-SLERP} emerges as the strongest model, achieving an average accuracy of $74.94\%$ and significantly outperforming all other models. Notably, \textit{DeFineMed-Qwen2.5-7B-SLERP} demonstrates exceptionally strong performance for its size, performing comparably to the much larger non-specialized \textit{Mistral-Small-24B-Instruct} model on the \textit{Anatomy} and \textit{Clinical Knowledge} benchmarks.

\subsection{Pairwise Win-Rate Evaluation}

This section presents the comparative performance of model checkpoints through pairwise win-rate analysis, quantified using the LLM-as-a-Judge methodology using \textit{GPT-4.1-mini} \citep{GPT-4.1-mini} on a machine-translated version of the \textit{MedAlpaca} dataset \citep{han2025medalpaca}. The evaluation prompts are provided in Appendix \ref{app:eval-prompts}. Given the requirement for instruction-following abilities, our analysis concentrates on the base instruction-tuned models and the final merged checkpoints. 

\begin{figure}
    \centering
    \includegraphics[width=\linewidth]{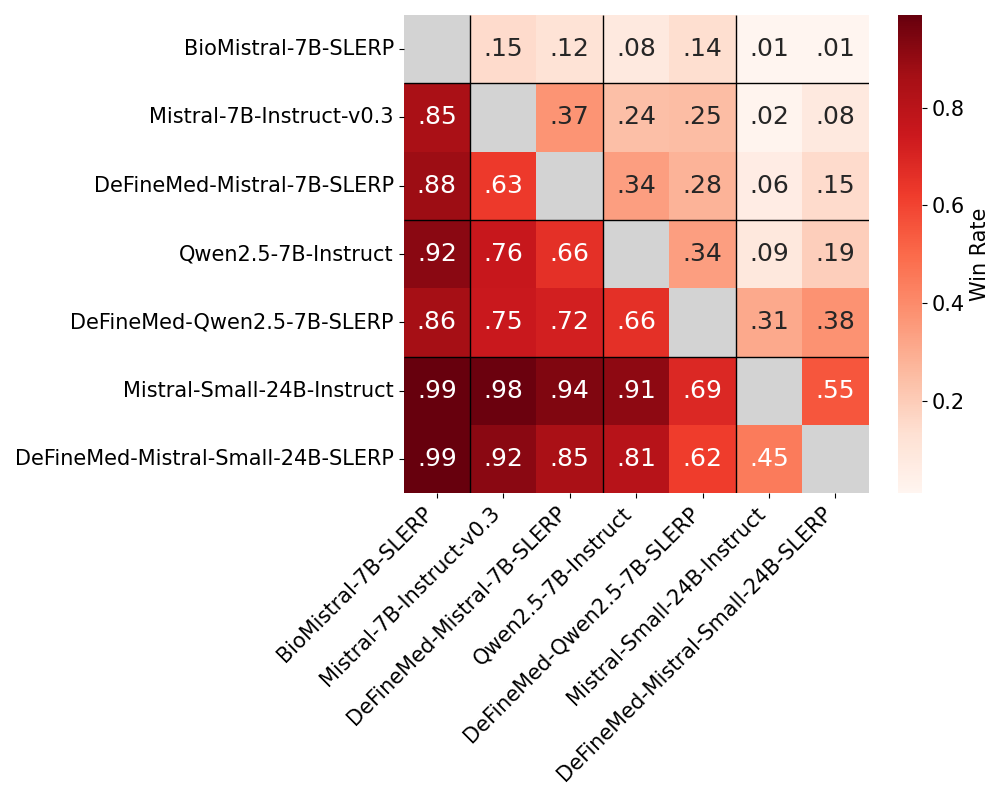}
    \caption{Matrix of model win-rates on the German \textit{MedAlpaca} dataset, where the value represents the win-rate of the row model over the column model in a head-to-head comparison.}
    \label{fig:pairwise_winrates}
\end{figure}

The results are summarized in Figure \ref{fig:pairwise_winrates}. The analysis confirms that model scale is the most significant determinant of output quality. The two $24B$ models overwhelmingly dominate all $7B$ models, establishing a high performance ceiling for the task.

Domain adaptation is highly successful in the $7B$ range. Pairwise comparisons show that the merged checkpoints consistently and significantly outperform their instruction-tuned base counterparts. For instance, \textit{DeFineMed-Qwen2.5-7B-SLERP} wins against \textit{Qwen2.5-7B-Instruct} by a rate of $0.66$ versus $0.33$. This validates model merging as a highly effective technique for applying domain knowledge while retaining core instruction capabilities at this scale. Among the $7B$ models, the \textit{Qwen2.5} architecture demonstrates superior base performance over the \textit{Mistral} architecture.

Crucially, specialization drastically closes the performance gap toward the largest model. The merged \textit{DeFineMed-Qwen2.5-7B-SLERP} model achieves a win-rate of $0.31$ against the much larger \textit{Mistral-Small-24B-Instruct}, representing a significant, $3.5$-fold increase from the base \textit{Qwen2.5-7B-Instruct} win-rate of $0.09$ against the same model. This demonstrates that the fusion of domain knowledge makes the smaller $7B$ model substantially more competitive against $24B$ models.

Conversely, the applied domain adaptation technique proved detrimental at the $24B$ scale, where the base \textit{Mistral-Small-24B-Instruct} model outperforms its merged counterpart.
This suggests that the performance drop of larger models may be partly attributable to limitations in the applied merging method, although additional factors related to large-scale adaptation cannot be excluded.

\subsection{Quantitative Evaluation of Failure Modes}

\begin{figure*}
    \centering
    \includegraphics[width=\linewidth]{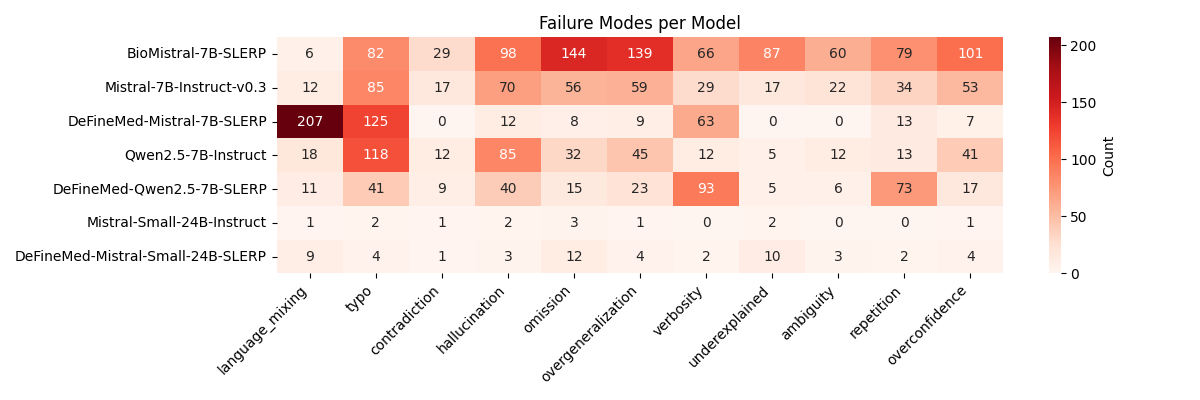}
    \caption{Frequency count of distinct failure modes for base instruction-tuned and merged models, quantified using \textit{GPT-4.1} methodology on $216$ instances from the German-translated \textit{MedAlpaca} dataset.}
    \label{fig:failure_modes}
\end{figure*}

This section presents a quantitative analysis of model output quality by analyzing failure modes. We again make use of the machine-translated version of the \textit{MedAlpaca} dataset \citep{han2025medalpaca} and employ \textit{GPT-4.1} to systematically quantify failure modes of LLM responses (see Appendix \ref{app:eval-prompts} for the prompt template).
Specifically, we analyze the frequency of failure modes including language mixing, typos, contradiction, hallucination and verbosity among others (see Table \ref{tab:failure_modes_descriptions} in the appendix for detailed descriptions).
Figure \ref{fig:failure_modes} summarizes the distribution of these failure modes across the selected model checkpoints. Overall, the results indicate that specialization via continual pre-training and subsequent model merging often mitigates a majority of common failure modes.

Across the \textit{Mistral-7B} and \textit{Qwen2.5-7B} families, a clear trend of mitigation is observed in failure modes such as hallucination, omission, overgeneralization, and repetition after the merging step. Furthermore, the integration of domain knowledge via merging significantly improves the models' epistemic calibration. Both overconfidence and contradiction errors are dramatically reduced. This suggests the specialized models are less likely to confidently state incorrect or conflicting information, leading to more factually grounded outputs.

Conversely, pre-training and merging can promote or introduce specific failure modes. The most extreme case is the language mixing failure mode in the \textit{DeFineMed-Mistral-7B-SLERP} model, which exhibited $207$ of the $216$ instances, a drastic increase from the $12$ instances in the base \textit{Mistral-7B-Instruct-v0.3}. This phenomenon aligns with prior qualitative observations and indicates a strong, yet localized, negative impact of multilingual domain adaptation on language separation. However, this trend is not observed in the \textit{Qwen2.5-7B} models.

We also observe an inverse relationship between omission and verbosity among the $7B$ models. The desire to be more comprehensive (reducing omission) often results in outputs that are significantly more verbose. This increase in verbosity is expected, as the models were continually pre-trained on a rich, verbose corpus of specialized text. The merging technique is not potent enough to fully counteract this learned behavior, suggesting that more involved instruction-tuning methods may be necessary to shape the desired output.

The occurrence of typos showed a contradictory pattern: a drastic decrease from $118$ to $41$ for the \textit{Qwen2.5-7B} family following specialization and merging, but a drastic increase from $85$ to $125$ for the \textit{Mistral-7B} family. This suggests that the impact on low-level language fluency is highly dependent on the underlying base model.

Finally, the $24B$ model family demonstrates remarkable stability. The failure mode counts for the merged checkpoint show no significant changes from its base model counterpart, with all counts remaining low. This observation is consistent with findings that larger models are more robust to failure modes introduced during post-pre-training adaptation, including perturbations from parameter merging or fine-tuning \citep{wei2022, yadav2024}.

\section{Discussion}
\label{sec:discussion}

A key element enabling our results is the construction of a large-scale German medical pre-training corpus from the \textit{FineWeb2 dataset}, achieved through a combination of LLM-based filtering and traditional ML techniques. This process directly addresses the limited availability of high-quality datasets and provides the necessary knowledge foundation for small model effectiveness.

The evidence from our evaluation directly addresses the question of whether small, domain-adapted models can challenge the performance of significantly larger counterparts. While a fully decisive answer requires further investigation, our findings represent a major step forward, demonstrating that the $7B$ models, enhanced via continual pre-training and merging, significantly closed the knowledge and performance gap against the $24B$ model.

In our quantitative benchmark analysis, the \textit{DeFineMed-Qwen2.5-7B-SLERP} model demonstrated performance comparable to the much larger \textit{Mistral-Small-24B-Instruct} on three out of four domain-specific benchmarks. More critically, the pairwise win-rate evaluation quantified this competitive gain: the specialization and merging process enabled the \textit{DeFineMed-Qwen2.5-7B-SLERP} to increase its win-rate against the \textit{Mistral-Small-24B-Instruct} by approximately $3.5$-fold. This marked reduction in the performance disparity supports the viability of adopting smaller, specialized models as a competitive, resource-efficient alternative to deploying large, general-purpose models for complex medical instruction-following tasks.

Notably, the magnitude of improvement differs between the two evaluation strategies, highlighting an important distinction in what each captures. While knowledge-intensive benchmarks measure absolute factual performance in a controlled, closed-ended setting, the pairwise win-rate evaluation additionally captures dimensions such as completeness, clarity, and practical helpfulness, reflecting real-world open-ended interactions more closely. The $7B$ \textit{Qwen2.5} model narrowed the average knowledge benchmark gap by nearly $2$-fold, confirming that domain adaptation can yield substantial gains across both evaluation paradigms.

Despite the successful performance gains, the specialization process highlighted critical trade-offs and limitations of the current techniques. The benchmark evaluation demonstrated that model merging, while successfully restoring instruction-following, yielded only marginal average improvements and inconsistent benefits, suggesting that merging alone is insufficient to fully harness the potential of domain-adapted checkpoints.

The quantitative analysis of failure modes further detailed these shortcomings. While specialization effectively mitigated factual errors like hallucination and overconfidence, it introduced mode-specific failures, most notably language mixing and increased verbosity due to the nature of pre-training. These findings underscore the inherent trade-off between domain adaptation and instruction-following fidelity. Furthermore, at the $24B$ scale, the merged model underperformed relative to the base model. This negative result may stem from several factors, including scale sensitivity of the SLERP merging method or hyperparameter configurations that were not specifically optimized for larger models. This highlights that CPT combined with SLERP merging is not guaranteed to improve performance at larger scales and underscores the need for scale-aware adaptation strategies. Collectively, these findings highlight that merging must be complemented by more targeted instruction-tuning techniques to fully harness the potential of specialized models.

\section{Conclusion}
\label{sec:conclusion}

In this paper, we successfully demonstrated a systematic approach to creating specialized German medical large language models, addressing the core question of whether small, resource-efficient models can be adapted to compete with significantly larger, general-purpose counterparts.

Our approach involved the creation of \textit{FineMed-de}, a large-scale German medical dataset, which was then used to successfully adapt three state-of-the-art LLMs ranging from $7B$ to $24B$ parameters. Our comprehensive evaluation confirmed the competitive viability of the $7B$ models. The pairwise win-rate analysis was particularly decisive, showing that domain adaptation and subsequent merging enabled the \textit{Qwen2.5}-based $7B$ model to achieve an approximately $3.5$-fold increase in its win-rate against the \textit{Mistral-Small-24B-Instruct} model. These results confirm that well-adapted smaller models represent a competitive and resource-efficient alternative for environments constrained by computation and regulation.

However, the evaluation also revealed critical trade-offs. While model merging successfully restored instruction-following capabilities, our failure mode analysis highlighted side effects such as language mixing and increased verbosity. These findings suggest that merging alone is insufficient to fully realize the potential of domain-adapted models. Moving forward, future research should prioritize domain-specific instruction tuning to mitigate the observed failure modes.

\section*{Limitations}

Despite the advancements achieved through continual pre-training and model merging, several limitations must be considered when interpreting the results.

One key limitation is the pre-training corpus, which was derived from \textit{FineWeb2} by filtering for German medical content. While this approach allowed us to construct a domain-specific dataset tailored to our needs, it also means that the dataset inherits the limitations of \textit{FineWeb2}. These include potential biases, misinformation, and the presence of harmful content. Notably, these issues persist despite the existing filtering efforts of the \textit{FineWeb2} authors, highlighting the challenges of fully eliminating such limitations. Consequently, our work inherits these limitations.

Another challenge lies in the process of model merging. While merging helped restore the instruction-following abilities that were lost during continual pre-training, its benefits were not consistent across different tasks. Moreover, the merging process introduced new failure modes. These unintended side effects underscore the need for further refinement after model merging to ensure more reliable and predictable improvements in model performance.

In addition, our evaluation primarily utilized the German subsets of two established benchmarks: \textit{MMMLU} and \textit{MedQA}. However, certain components, such as the \textit{College Medicine} task in \textit{MMMLU}, are known to contain a significant proportion of incorrectly labeled samples \citep{gema2025mmlu}. Furthermore, a key limitation of this study is our reliance on machine-translated benchmarks, due to the lack of publicly available, expert-curated German medical QA datasets, a common challenge in multilingual LLM evaluation \citep{de-vroe-etal-2025-comparing}.

While the pairwise win-rate and failure mode analyses provide valuable insights into relative model behavior and output quality, they also entail several limitations. 
First, the LLM-as-a-Judge evaluation relies on \textit{GPT-4.1-mini} as the scoring model, which introduces potential bias and may not perfectly align with human judgment, particularly for multilingual or domain-specific content. Furthermore, pairwise comparisons capture relative preference, not absolute accuracy, and may favor stylistic similarity to the judge model over true factual correctness. 
Second, the failure mode quantification depends on automated classification of response types, which may not fully capture nuanced language or reasoning mistakes. 
The analysis is also limited by the scope of the \textit{MedAlpaca} dataset, which may not capture the full range of medical or general instruction-following scenarios. Future work should include human expert evaluations, cross-judge consistency checks, and domain-specific annotation frameworks to validate and extend these findings.

\section*{Acknowledgments}

This work was done within the project SmartHospital.NRW with grant number 005-2011-0041/2 and project number 2011ki001b, funded by the Ministry for Economic Affairs, Industry, Climate Action and Energy of the State of North Rhine-Westphalia, Germany. We thank the reviewers for their valuable feedback.

\bibliography{custom}

\newpage

\appendix
\section{Appendix}

\subsection{Medical Document Pre-Classifier}
\label{sec:med_pre_cls}

To classify documents into medical and non-medical categories, we apply the \textit{Mixtral-8x7B-Instruct-v0.1} model with a zero-shot prompting approach. The exact prompt used for classification is as follows:

\begin{quote}
You are a medical expert tasked with identifying whether the provided
content is both medical and high quality. Analyze the content carefully based
on the following criteria: \\
- The content is related to health, medicine, or healthcare. \\
- The information needs to be scientific, high-quality, accurate, and well-structured. \\
Only allow the highest quality data. If it doesn't seem scientific, then leave it out. Also,
leave out news stories or similar styles of text that are not written by medical experts.
Restrict your response to 'yes' or 'no'. Do not include any other explanation. \\
--- \\
Content: <DOCUMENT>
\end{quote}
This prompt is used to filter a subset of approximately $260k$ documents from the German \textit{FineWeb2} dataset. The model is constrained to generate only a single token, a document is classified as medical if the output token is "yes"; otherwise, it is categorized as non-medical.

\subsubsection{Performance Evaluation of Medical Document Pre-Classifier}

To evaluate the performance of the LLM-based medical document classifier, we randomly sampled $100$ documents from the dataset, consisting of $50$ documents classified as medical and $50$ classified as non-medical by the LLM. These were annotated by three human annotators, who independently assessed the class of each document. 
We achieved an inter-annotator agreement of $84.7\%$, measured in \textit{Fleiss' Kappa} \citep{randolph2005free}.

The evaluation results, presented in Table \ref{tab:llm_scores}, demonstrate a strong overall performance of the LLM-based classifier, with an F1-score of $91.1 \pm 2.5$. The overall performance suggests that the LLM effectively distinguishes between medical and non-medical documents, although minor discrepancies with human annotators were observed. To more closely evaluate the disagreement between the LLM and human annotators, we inspect the false positives in Table \ref{tab:llm_false_positives}. We find that, while some false positives are clearly non-medical, others suggest a degree of relevance to the medical domain.

Given these findings, we consider the LLM's performance sufficiently robust to proceed with training a more specialized classifier using the resulting labeled \textit{FineMed-de-260k} dataset.

\begin{table}[t]
\centering
\begin{tabular}{l|c}
\textbf{Metric} & \textbf{Score} \\
\hline
Precision & $92.0 \pm 2.9$ \\
Recall &  $90.3 \pm 2.6$ \\
F1-score & $91.1 \pm 2.5$ \\
Accuracy & $91.0 \pm 2.4$ \\
\end{tabular}
\caption{Evaluation scores of the LLM's performance against three human annotators on a random subset of $100$ samples.}
\label{tab:llm_scores}
\end{table}

\begin{table*}[h]
\centering
\begin{tabular}{|c|p{0.85\linewidth}|}
\hline
\textbf{Count} & \textbf{False Positive Example} \\
\hline
3 & Siehe auch: Mango Chicken Curry – Eine köstliche und gesunde Mahlzeit. Inhalt: 1. Erdmandeln sind reich an Ballaststoffen, 2. Erdmandeln sind reich an Vitaminen und Mineralstoffen, 3. Erdmandeln sind glutenfrei, 4. Erdmandeln sind ... \\
\hline
2 & Es gibt sie wie Sand am Meer – die Persönlichkeitstest. Wir fragen uns, inwiefern ein solcher Test wirklich zu einem persönlichen Turnaround führen kann. Marlies Zindel führt eine eigene Beratungspraxis und nutzt den sogenannten Birkman-Test in ihren Gespr... \\
\hline
2 & Dr. Preis ist unter den Top5:· Orthopäden in Köln (Stand 06/2017) Dr. med. Stefan Preis Arzt, Orthopäde Weiterbildungen: Chirotherapie (Manuelle Medizin), Sportmedizin Orthopädie/Sporttraumatologie Klinik am Ring Hohenstaufenring 28 50674 Köln ... \\
\hline
1 & Video: Was versteht man unter dem Begriff Osmolarität? Osmotische Konzentration, früher bekannt als Osmolarität, ist das Maß für die Konzentration des gelösten Stoffes, definiert als die Anzahl der Osmole (Osm) des gelösten Stoffes pro Liter (L) der Lösung... \\
\hline
1 & Viele Menschen haben den Verdacht, dass Tilapia "schlecht" oder "schmutzig" sei, was viele zu der Frage führt, ob der Fisch schlecht für Sie ist. Das liegt daran, dass Tilapia für seine Kontamination bekannt ist. In der Vergangenheit ernährten sich einige... \\
\hline
1 & Forschungsförderung für UMG-Ernährungsprojekt Risiken früher erkennen, Komplikationen verhindern Forschende der Unimedizin Greifswald wollen die Überlebenschancen von Patienten mit entzündeter Bauchspeicheldrüse steigern. Dazu möchten sie den Einfluss des... \\
\hline
1 & Anders als beim Sehen, können wir nur schwer störende Geräusche ausblenden. Deswegen wirkt sich unerwünschter Lärm auch auf unseren Schlafqualität aus, erklärt Prof. Dr. med. Dipl.-Psych. Manfred Beutel. Nicht nur physiologische, sondern auch seelische... \\
\hline
\end{tabular}
\caption{Excerpt of the false-positives from the manually annotated 100 samples, showing the text excerpts along with the number of annotators who deemed each sample non-medical.}
\label{tab:llm_false_positives}
\end{table*}

\begin{table}[]
    \centering
    \begin{tabular}{l|c}
        \textbf{Parameter} & \textbf{PT} \\
        \hline
        Learning Rate & $2 \times 10^{-5}$ \\
        Weight Decay & $0.01$ \\
        Warmup Steps & 500 \\
        Effective Batch Size & $1024$ \\
        Sequence Length & $8192$ \\
    \end{tabular}
    \caption{Hyperparameters used for continual pre-training (PT).}
    \label{tab:hyperparameters}
\end{table}

\subsection{Training of Medical Classifier}\label{app:med_cls}

To efficiently classify documents as medical or non-medical, we fine-tune a \textit{XLM-RoBERTa-base} model using the labeled \textit{FineMed-de-260k} dataset. This fine-tuning process aims to optimize the model for high precision in distinguishing medical content. The model was trained for $5k$ steps, approximately two epochs, with a batch size of $96$, utilizing the \textit{AdamW} \citep{loshchilov2019decoupledweightdecayregularization} optimizer and a learning rate of $2 \times 10^{-5}$, with a weight decay of $0.01$. Given the emphasis on precision over recall, we used the $\textrm{F}_{\beta}$ score with $\beta = 0.7$ as the early stopping criterion.

\begin{figure}[h]
    \centering
    \includegraphics[width=.9\linewidth]{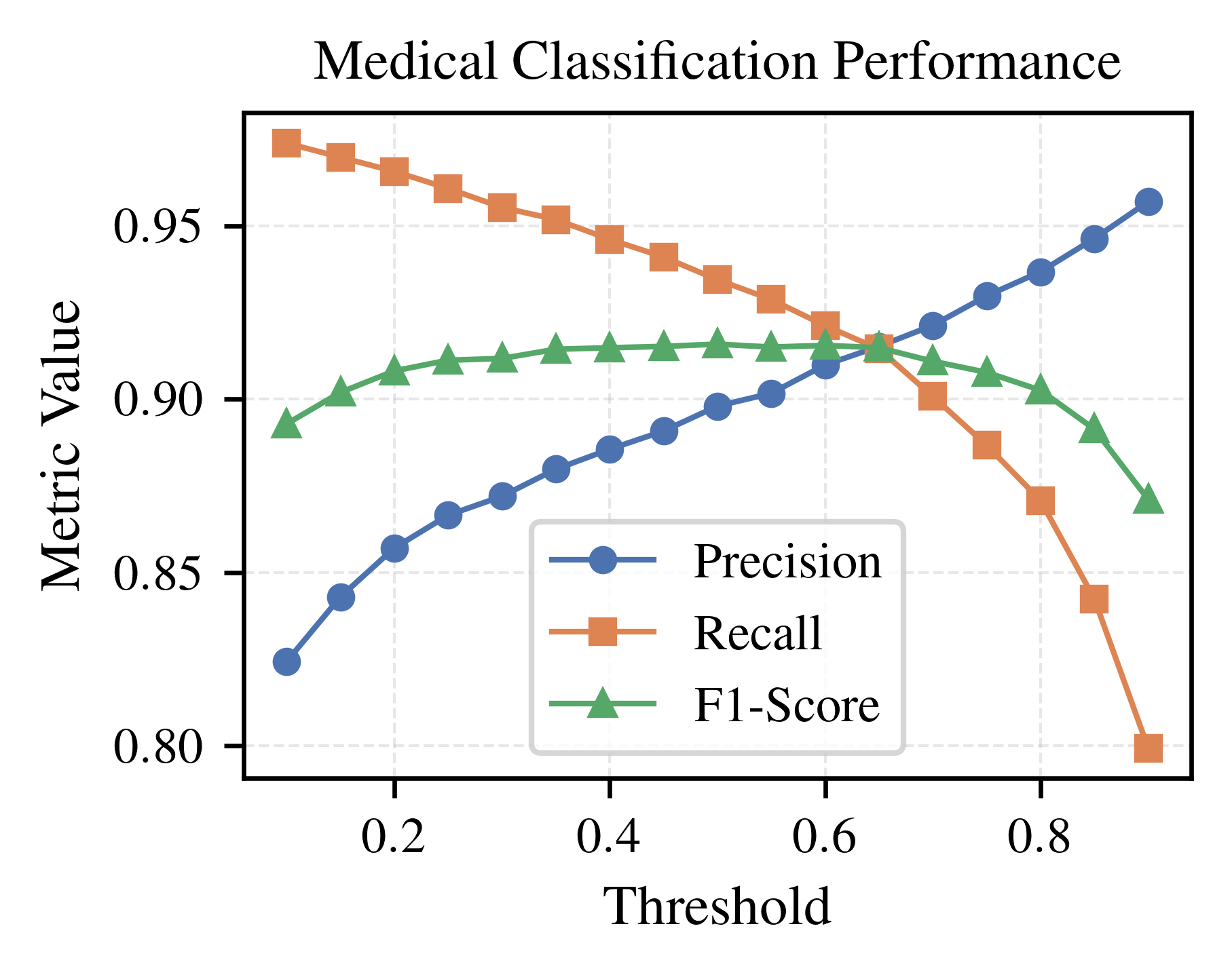}
    \caption{Performance metrics of the medical document classifier for various decision thresholds.}
    \label{fig:classifier_performance}
\end{figure}

To ensure high precision during filtering, we employ a decision threshold of $0.9$. This threshold is strategically chosen to minimize the inclusion of non-medical texts, prioritizing the precision of the resulting medical document selection. As depicted in Figure \ref{fig:classifier_performance}, this approach yields a precision of $0.95$ and a recall of $0.8$ at the chosen threshold of $0.9$, demonstrating an effective balance between these metrics. This application process allows for the reliable extraction of medical documents, forming the foundation of our specialized dataset.

\begin{figure}[t]
\centering
\includegraphics[width=.8\linewidth]{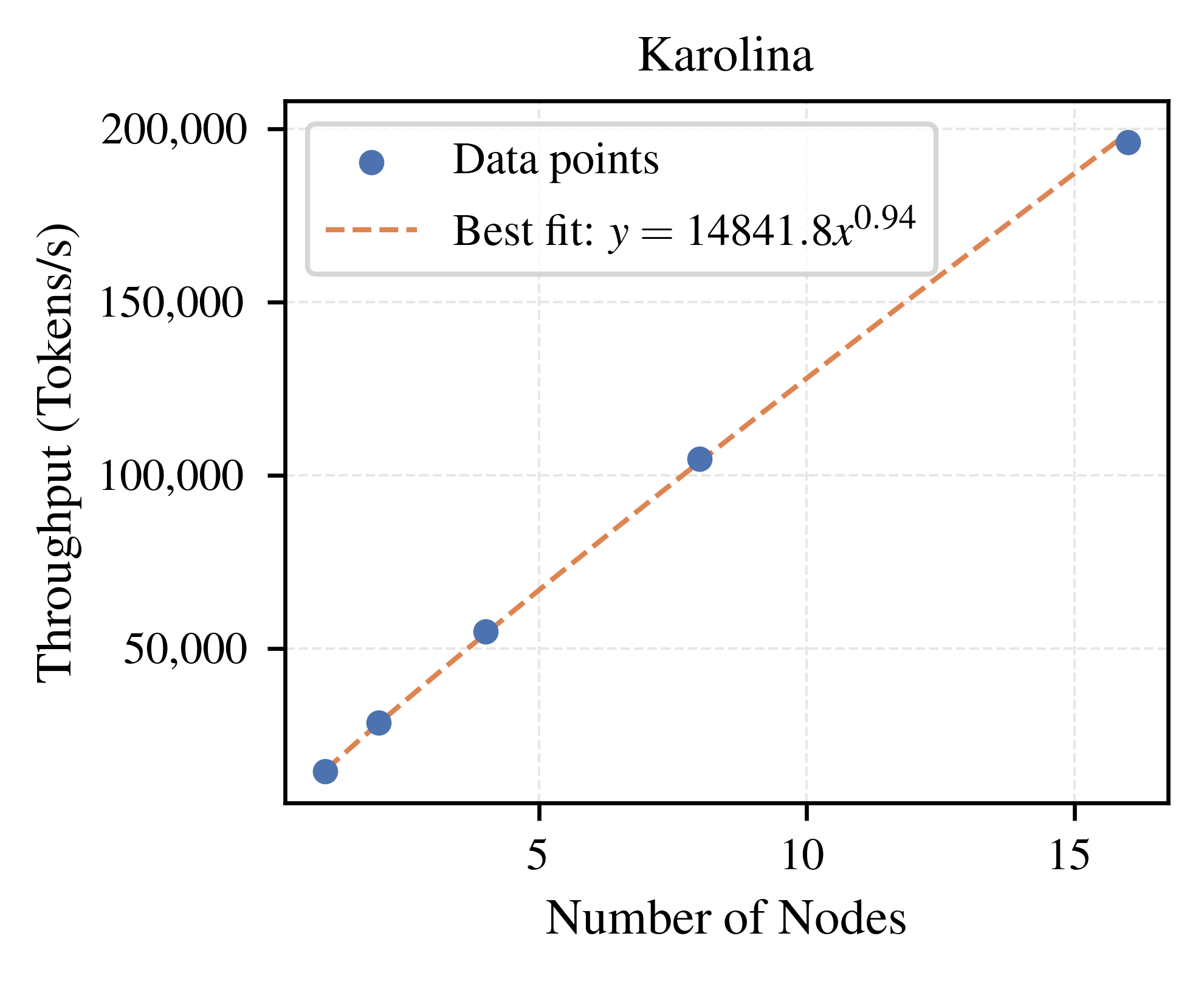}
\includegraphics[width=.8\linewidth]{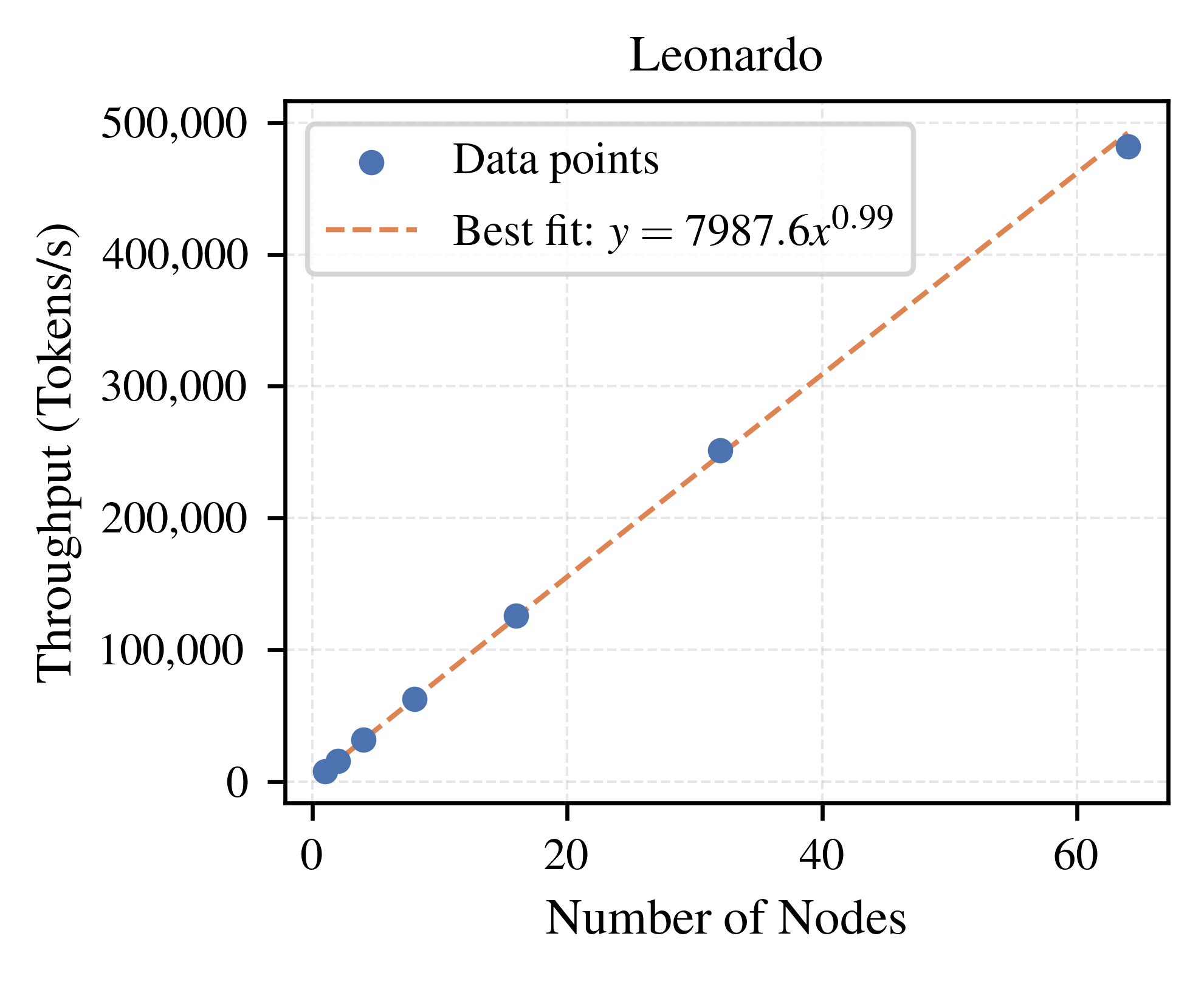}
\caption{Weak scaling behavior on Karolina and Leonardo. The actual computation per accelerator is kept constant throughout, with a micro batch size of 1.}
\label{fig:scaling}
\end{figure}

\subsection{Continual Pre-Training Details}
\label{app:training-details}

The following outlines specific technical details involved in the continual pre-training process for our models. We describe the data handling strategies employed for efficient batching, the optimization setup and training schedule used to ensure stable and effective learning, and the hardware resources and distributed training techniques leveraged to scale our experiments.

For batching efficiency, we adopt a bin-packing strategy, grouping sequences into constant-length batches to reduce padding and truncation. This approach not only avoids the computational overhead of padding tokens but also results in better model performance and higher throughput \citep{ding2024fewer}. We separate different sequences within a sample using the tokenizer's \texttt{<EOS>} token, allowing them to attend to each other within the same batch. This method contributed significantly to the overall efficiency of the pre-training process.

We train the models using the \textit{AdamW} optimizer \citep{loshchilov2019decoupledweightdecayregularization} with a linear learning rate decay. In order to address training instabilities, we implement an extended warmup phase of $500$ steps, representing about half of the medical training corpus. To compensate for the longer warmup phase, we train all models for two epochs. Research by \citet{tunstall2023zephyrdirectdistillationlm} suggests that training beyond $1.5$ epochs yields minimal additional benefits, which supports our training regime of 2 epochs with $0.5$ epochs of warmup. Additionally, \citet{muennighoff2023scaling} indicate that four epochs are typically necessary before overfitting becomes a concern, further justifying our decision.

The experiments were conducted across multiple compute clusters. For most models, we utilized 8 nodes on the Karolina cluster, each equipped with 8 NVIDIA A100 (40GB) GPUs. However, the Mistral-Small-24B model was trained on the Leonardo cluster\footnote{\url{https://leonardo-supercomputer.cineca.eu/}}, leveraging 128 nodes. Table \ref{tab:compute_efforts} provides a summary of the hardware and compute resources used during the continual pre-training phase for each model. Although all models were trained for exactly two epochs, the number of steps per model varied. This discrepancy is due to differences in the model-specific tokenizers and their fertility on the training corpus. Detailed hyperparameters are listed in Table \ref{tab:hyperparameters}.

For multi-node settings, we employed the Hybrid Sharding Data Parallel (HSDP) variant of FSDP. This approach involved distributing the data across multiple compute nodes and sharding the model parameters, gradients, and optimizer states within each node. By avoiding node-level sharding, we achieved near-linear scaling, as illustrated in Figure \ref{fig:scaling}. In both cases, we reached a throughput of approximately 1850 tokens per second per GPU for $7B$ parameter sized models. This is consistent with findings of prior works, which report throughputs of 2350 \citep{llama3_german_8b} and 1950 \citep{labrak2024biomistralcollectionopensourcepretrained} tokens per second per GPU, respectively. Note that, due to its larger size, the \textit{Mistral-Small-24B} model required full sharding across nodes, leading to a corresponding increase in communication overhead.

The NVIDIA A100 accelerators available to us were equipped with only 40GB (Karolina) and 64GB (Leonardo) of VRAM. As a result, we had to reduce our micro batch size to 1, which lowered the arithmetic intensity compared to what would be possible on the more typical 80GB A100 GPUs.

\begin{table*}[t]
    \centering
    \resizebox{\textwidth}{!}{
        \begin{tabular}{l|c|c|c|c|c|c}
            \toprule
            \textbf{Model} & \textbf{Cluster} & \textbf{Accelerator} & \textbf{Num GPUs} & \textbf{Time} & \textbf{Steps} & \textbf{Tokens} \\
            \midrule
            Mistral-7B-Instruct-v0.3 & Karolina & A100 (40GB) & 64 & 72h & 2970 & 25.1B \\
            Qwen2.5-7B-Instruct & Karolina & A100 (40GB) & 64 & 64h & 2634 & 22.1B \\
            \midrule
            Mistral-Small-24B-Instruct & Leonardo & A100 (64GB) & 128 & 48h & 2294 & 19.2B \\
            \bottomrule
        \end{tabular}
    }
    \caption{Computational resources invested in continual pre-training of different models.}
    \label{tab:compute_efforts}
\end{table*}

\begin{table*}[t]
    \centering
    \resizebox{\textwidth}{!}{
    \begin{tabular}{l|cccc|c}
    \textbf{Model} & \textbf{Anatomy} & \textbf{Clinical Knowledge} & \textbf{College Medicine} & \textbf{MedQA} & \textbf{Average} \\ 
    \hline
	Qwen2.5-7B-Instruct &54.07 ± 4.30 &69.06 ± 2.85 &63.01 ± 3.68 &50.20 ± 2.24 &59.08 ± 1.68 \\ 
    \hline
    DeFineMed-Qwen2.5-7B &\underline{62.22 ± 4.19} &\underline{76.60 ± 2.61} &\underline{68.21 ± 3.55} &52.60 ± 2.24 & \underline{64.91 ± 1.62} \\ 
	FineWeb2-Qwen2.5-7B &56.30 ± 4.28 &75.47 ± 2.65 &61.85 ± 3.70 & \underline{53.20 ± 2.23} &61.70 ± 1.66 \\
    \hline
    DeFineMed-Qwen2.5-7B-SLERP &\underline{65.19 ± 4.12} &\underline{75.85 ± 2.63} &65.90 ± 3.61 &54.20 ± 2.23 &\underline{65.28 ± 1.62} \\ 
	FineWeb2-Qwen2.5-7B-SLERP &55.56 ± 4.29 &73.58 ± 2.71 &\underline{68.21 ± 3.55} &\underline{54.40 ± 2.23} &62.94 ± 1.65 \\ 
    \hline
    \end{tabular}
    }
    \caption{Performance comparison of Qwen2.5-7B-Instruct models trained on domain-specific \textit{FineMed-de} and randomly sampled \textit{FineWeb2-de} data. Best score is \underline{underlined}.}
    \label{tab:ablation_model_performance}
\end{table*}

\subsection{Merge Configuration}
\label{app:merge-config}

We provide the SLERP merge configuration used for all models, following \citet{Lu2025finetuning}. Rather than a uniform interpolation weight, we apply component-specific gradient schedules across the model depth for self-attention and MLP layers, while remaining parameters use $t=0.5$.

\begin{verbatim}
slices:
  - sources:
      - model: <base-model>
        layer_range: [0, <num_layers>]
      - model: <cpt-model>
        layer_range: [0, <num_layers>]
merge_method: slerp
base_model: <base-model>
parameters:
  t:
    - filter: self_attn
      value: [0, 0.5, 0.3, 0.7, 1]
    - filter: mlp
      value: [1, 0.5, 0.7, 0.3, 0]
    - value: 0.5
dtype: float16
\end{verbatim}

\subsection{Impact of Domain-Specific Filtering}
\label{sec:ablation-domain-specific-filtering}

 To rigorously assess the impact of utilizing a highly domain-specific dataset, we conduct an ablation study designed to isolate and quantify the benefit of our medical corpus compared to a baseline of unfiltered data. To this end, we create a subset of the \textit{FineWeb2} dataset, matching the size of our filtered medical corpus at $10.0B$ tokens. Applying the medical document classifier from Section \ref{sec:classifier}, reveals that only $0.89\%$ of the sampled documents are related to the medical domain. We continually pre-trained the \textit{Qwen2.5-7B-Instruct} model, using the exact same hyperparameters as discussed in Section \ref{sec:pretraining}, for two epochs, resulting in a total of $22.1B$ training tokens and evaluate under the same conditions as the other models.

The performance results, presented in Table \ref{tab:ablation_model_performance}, demonstrate that the model trained on the random sample shows improvements over its base model. Interestingly, we find that the model trained on the random subset outperforms its domain-specific counterparts in three out of eight settings. This especially includes the \textit{MedQA} benchmark. These exceptions need further investigation. On average however, it is clearly outperformed by the domain-specific model variants. The improvements observed in \textit{FineWeb2-Qwen2.5-7B-Instruct} over the base model can be attributed to the inherent high quality of the \textit{FineWeb2} dataset and the general benefit of continual pre-training on German data. However, the substantially larger gains achieved with the model trained with the filtered medical corpus demonstrate the significant impact of a domain-specific corpus.

\subsection{Evaluation on Non-medical tasks}
\label{sec:non_medical_eval}

To investigate potential trade-offs in instruction-following capabilities introduced by our domain-specific adaptation, we systematically evaluated the \textit{DeFineMed} models on non-medical benchmarks. In particular, we compared the performance of the original pre-trained baseline models with the versions obtained after continual pre-training on medical data and subsequent model merging. 
For this evaluation, we selected several widely used benchmarks outside the medical domain: \textit{MMMLU} \citep{hendryckstest2021, hendrycks2021ethics}, which measures broad multi-task generalization across knowledge areas; \textit{ARC-Easy} and \textit{ARC-Challenge} \citep{clark2018arc} for scientific and commonsense question answering; \textit{GSM8k} \citep{cobbe2021gsm8k} to assess mathematical reasoning; and \textit{HellaSwag} \citep{zellers2019hellaswag} for evaluating contextual understanding and narrative completion. This setup allows us to quantify how well the adapted models retain general-purpose reasoning and instruction-following skills beyond their specialized medical context. As shown in Table \ref{tab:non_medical_performance}, the medical models consistently outperform their base versions, suggesting that continual pre-training on our dataset not only enhances domain alignment on medical benchmarks but also improves general linguistic competence. 

\subsection{Evaluation Prompts}
\label{app:eval-prompts}

This section provides the prompt templates used in the pairwise win-rate evaluation (Section \ref{sec:evaluation}) and the failure mode analysis. All templates use Jinja2 syntax for variable substitution.

\subsubsection{Response Generation}

For both the pairwise win-rate and failure mode evaluations, each model is presented with the query from the German-translated \textit{MedAlpaca} dataset using the chat template of its corresponding base instruction-tuned model. No additional system prompt is applied.

\subsubsection{Pairwise Win-Rate Prompt}

The following prompt template is used with \textit{GPT-4.1-mini} for the LLM-as-a-Judge pairwise comparison. The model is instructed to return a structured JSON response.

{\small
\begin{verbatim}
[System]
You are an AI assistant that evaluates pairs of
responses to a given query. Your goal is to
determine which response is better based on
correctness, clarity, completeness, and relevance.

[User]
Compare the following two responses. Your answer
needs to follow this json-schema:

```json
{
  "type": "object",
  "properties": {
    "reasoning": {
      "type": "string",
      "description": "Detailed explanation of why
        this decision was made, analyzing the
        quality and characteristics of both
        responses."
    },
    "winner": {
      "type": "string",
      "enum": ["A", "B", "0"],
      "description": "The winning response, i.e.
        'A', 'B' or '0' indicating a tie."
    }
  },
  "required": ["reasoning", "winner"]
}
```

Context:
  {{ context }}

Query:
  {{ query }}

Response A:
  {{ response_a }}

Response B:
  {{ response_b }}

Based on the query and context above, which
response is better?
\end{verbatim}
}

\subsubsection{Failure Mode Analysis Prompt}

The following prompt template is used with \textit{GPT-4.1} to classify failure modes in model responses. The model evaluates each failure mode independently and returns a structured JSON response. See table \ref{tab:failure_modes_descriptions} for a complete list of all failure modes and their descriptions.

{\small
\begin{verbatim}
[System]
You are an evaluator of language model responses.

Your task is to analyze whether the given response
exhibits any of the following failure modes.

```json
{
  "type": "object",
  "properties": {
    "language_mixing": {
      "type": "boolean",
      "description": "Whether the response
        unexpectedly switches between languages
        or dialects without instruction or user
        prompting."
    },
    "typo": {
      "type": "boolean",
      "description": "Whether the response
        contains spelling mistakes, character
        errors, or malformed words that reduce
        readability."
    },
    ...
  },
  "required": [
    "language_mixing", "typo", "contradiction",
    "hallucination", "omission",
    "overgeneralization", "irrelevance",
    "verbosity", "underexplained", "ambiguity",
    "instruction_ignoring", "repetition",
    "overconfidence", "bias_or_harm"
  ]
}
```

Instructions:
1. Carefully read the translated question (the
   original user query) and the response (the
   model's answer).
2. For each failure mode, decide whether it
   applies to the response.
   - Answer true if the failure mode is present.
   - Answer false if the failure mode is not
     present.
3. Return your answer as valid JSON following the
   json-schema above.

[User]
Context:
  {{ context }}

Query (the question posed to the model):
<<<
  {{ query }}
>>>

Response (the model's answer to be evaluated):
<<<
  {{ response }}
>>>
\end{verbatim}
}

\begin{table*}[t]
\centering
\resizebox{\textwidth}{!}{
\begin{tabular}{l|p{0.75\textwidth}}
\toprule
\textbf{Failure Mode} & \textbf{Description} \\
\midrule
\textbf{Language Mixing} & Whether the response unexpectedly switches between languages or dialects without instruction or user prompting. \\
\textbf{Typo} & Whether the response contains spelling mistakes, character errors, or malformed words that reduce readability. \\
\textbf{Contradiction} & Whether the response contains internal inconsistencies or conflicts with its own earlier statements. \\
\textbf{Hallucination} & Whether the response fabricates facts, citations, or details that are presented as truth but are not grounded in reliable sources. \\
\textbf{Omission} & Whether key details, steps, or context are missing, leading to an incomplete or misleading answer. \\
\textbf{Overgeneralization} & Whether the response makes broad, unsupported claims or ignores edge cases, resulting in oversimplification. \\
\textbf{Verbosity} & Whether the response is unnecessarily long-winded, repeating points without adding value. \\
\textbf{Underexplained} & Whether the response is too brief or lacks the necessary depth to be useful or correct. \\
\textbf{Ambiguity} & Whether the response is vague or unclear in wording, leading to multiple possible interpretations. \\
\textbf{Repetition} & Whether the response repeats phrases or sentences excessively, often due to generation loops. \\
\textbf{Overconfidence} & Whether the response states information in an authoritative tone despite being incorrect or uncertain. \\
\bottomrule
\end{tabular}
}
\caption{Descriptions of Failure Modes}
\label{tab:failure_modes_descriptions}
\end{table*}

\begin{table*}[t]
    \centering
    \resizebox{\textwidth}{!}{
    \begin{tabular}{l|ccccc|c}
        \textbf{Model} & \textbf{MMLU} & \textbf{ARC-C} & \textbf{ARC-E} & \textbf{GSM8k} & \textbf{HellaSwag} & \textbf{Average}\\
        \hline
    	BioMistral-7B-SLERP (baseline) &46.87 ± 0.41 &40.96 ± 1.44 &63.43 ± 0.99 &2.35 ± 0.42 &54.00 ± 0.50 &41.52 ± 0.38 \\ 
        \hline
    	Mistral-7B-Instruct &50.33 ± 0.40 &48.29 ± 1.46 &68.81 ± 0.95 &11.22 ± 0.87 &59.82 ± 0.49 &47.70 ± 0.41 \\ 
    	Mistral-7B-Instruct-Medical &51.95 ± 0.40 &49.23 ± 1.46 &\underline{71.59 ± 0.93} &11.60 ± 0.88 &\underline{66.59 ± 0.47} &50.19 ± 0.41 \\ 
    	Mistral-7B-Instruct-Medical-SLERP &\underline{53.39 ± 0.40} &\underline{49.83 ± 1.46} &\underline{71.59 ± 0.93} &\underline{13.27 ± 0.93} &66.08 ± 0.47 &\underline{50.83 ± 0.41} \\ 
        \hline
    	Qwen2.5-7B-Instruct &63.35 ± 0.39 &51.28 ± 1.46 &68.81 ± 0.95 &58.23 ± 1.36 &61.05 ± 0.49 &60.54 ± 0.46 \\ 
    	Qwen2.5-7B-Instruct-Medical &64.44 ± 0.38 &51.28 ± 1.46 &\underline{70.71 ± 0.93} &63.38 ± 1.33 &\underline{63.22 ± 0.48} &62.61 ± 0.45 \\ 
    	Qwen2.5-7B-Instruct-Medical-SLERP &\underline{65.57 ± 0.38} &\underline{52.22 ± 1.46} &70.12 ± 0.94 &\underline{66.94 ± 1.30} &62.22 ± 0.49 &\underline{63.42 ± 0.45} \\ 
         \hline
    	Mistral-Small-24B-Instruct &71.95 ± 0.35 &64.08 ± 1.40 &80.98 ± 0.81 &\underline{64.97 ± 1.31} &71.96 ± 0.45 &\underline{70.79 ± 0.43} \\ 
    	Mistral-Small-24B-Instruct-Medical &71.26 ± 0.36 &62.97 ± 1.41 &80.89 ± 0.81 &53.75 ± 1.37 &73.15 ± 0.44 &68.41 ± 0.44 \\ 
    	Mistral-Small-24B-Instruct-Medical-SLERP &\underline{73.31 ± 0.35} &\underline{65.27 ± 1.39} &\underline{81.40 ± 0.80} &56.86 ± 1.36 &\underline{73.69 ± 0.44} &70.11 ± 0.44 \\ 
         \hline
    \end{tabular}
    }
    \caption{Model performance on different German non-medical benchmarks in one-shot setting. The average accuracy is the unweighted mean of accuracies, whereas the average standard error is the square root of the unweighted mean of variances. Best score is \underline{underlined}.}
    \label{tab:non_medical_performance}
\end{table*}

\end{document}